\documentclass{styles/svproc}
\usepackage{url}

\usepackage{hyperref}
\usepackage{marvosym}
\usepackage[usenames,dvipsnames]{xcolor}
\definecolor{floor}{RGB}{68,1,84}
\usepackage{amsfonts}
\usepackage{amsmath}
\usepackage{graphicx}
\usepackage{tabularx,booktabs}
\usepackage{xcolor}

\usepackage[a4paper,left=4.69cm,right=3.99cm,top=3.21cm,bottom=6.21cm]{geometry}

\begin{document}
\mainmatter              

\title{MonoNav: MAV Navigation via Monocular \\ Depth Estimation and Reconstruction}

\titlerunning{MonoNav}  
%
\author{Nathaniel Simon\textsuperscript{(\Letter)} \and Anirudha Majumdar}
\authorrunning{Nathaniel Simon et al.} 
%
\tocauthor{Nathaniel Simon and Anirudha Majumdar}
\institute{Department of Mechanical and Aerospace Engineering \\ Princeton University, Princeton NJ 08544, USA\\
\email{nsimon@princeton.edu} \thanks{Project Website: \href{https://natesimon.github.io/mononav/}{\texttt{natesimon.github.io/mononav}}. As seen at \href{https://iser2023.org/}{ISER 2023}.}}

\maketitle              
\begin{abstract}
A major challenge in deploying the smallest of Micro Aerial Vehicle (MAV) platforms ($\le$ 100~g) is their inability to carry sensors that provide high-resolution metric depth information (e.g., LiDAR or stereo cameras). Current systems rely on  end-to-end learning or heuristic approaches that directly map images to control inputs, and struggle to fly fast in unknown environments. In this work, we ask the following question: using \emph{only} a monocular camera, optical odometry, and offboard computation, can we create metrically accurate maps to leverage the powerful path planning and navigation approaches employed by larger state-of-the-art robotic systems to achieve robust autonomy in unknown environments? 
We present MonoNav: a fast 3D reconstruction and navigation stack for MAVs that leverages recent advances in depth prediction neural networks to enable metrically accurate 3D scene reconstruction from a stream of monocular images and poses. MonoNav uses off-the-shelf pre-trained monocular depth estimation and fusion techniques to construct a map, then searches over motion primitives to plan a collision-free trajectory to the goal. In extensive hardware experiments, we demonstrate how MonoNav enables the Crazyflie (a 37~g MAV) to navigate fast (0.5~m/s) in cluttered indoor environments. We evaluate MonoNav against a state-of-the-art end-to-end approach, and find that the collision rate in navigation is significantly reduced (by a factor of 4). This increased safety comes at the cost of conservatism in terms of a 22\% reduction in goal completion.

\keywords{MAV, monocular depth estimation, 3D reconstruction, collision avoidance}
\end{abstract}

\section{Introduction and Related Work}
The smallest class of unmanned aerial vehicles (UAVs), referred to as micro aerial vehicles (or MAVs, $\le$ 100~g), are well-suited for constrained indoor applications such as inspection, exploration, and mapping. However, their small size and weight restricts their ability to carry sensors that provide high-resolution metric depth information (e.g., LiDAR or stereo cameras). While there have been advances in sensor and computer miniaturization towards fully onboard systems, such as a 2.28~g Time of Flight depth sensor \cite{niculescu2022towards} or 4.4~g GAP8 embedded processor \cite{palossi2019open}, the low sensor resolution and limited compute result in low levels of autonomy (e.g., object avoidance within highly structured settings). 

Instead, our system assumes fully onboard sensing, but offboard computation --- either in the form of a nearby desktop, trailing ground vehicle, or cloud. We assume our MAV has access to the typical instrumentation: a forward-facing monocular camera, an inertial measurement unit (IMU) for orientation and acceleration, and an optical flow camera and height sensor for position and velocity estimation.

For such a setup, the prevailing approach to monocular, vision-based exploration is to train an end-to-end model which takes as input a color image and outputs a velocity setpoint or trajectory of waypoints.
In \cite{chaplot2020learning,chaplot2020neural}, a learned mapper model infers spatial relationships from an RGB image to construct a 2D overhead view for SLAM. This approach performs well in simulated environments, but lacks validation in the real world.
In \cite{sadeghi2016cad2rl}, a deep convolutional neural network trained entirely in simulation navigates a MAV through hallways in the real world. Such approaches, however, may overfit to the robots and environments seen in training.

To enable monocular navigation across diverse real-world environments, \cite{shah2023gnm} proposes a general-purpose, goal-image-based policy trained and deployed across a wide range of environments and robot embodiments. In addition to requiring sub-goal images,  \cite{shah2023gnm} requires a topological graph of the environment, which must be demonstrated (through teleoperation) or generated through cautious robot exploration \cite{shah2021rapid}. To overcome this and enable efficient exploration, ViNT \cite{shah2023vint} proposes sub-goal generation through \textit{image diffusion}; the sub-goal images are spatially grounded and scored by a heuristic (e.g., goal position). More recently, NoMaD \cite{sridhar2023nomad} trains a \textit{diffusion policy} \cite{chi2023diffusion}, which takes a series of images and outputs normalized action candidates. These action candidates are un-normalized into position trajectories using a robot-specific range. In a variety of environments, NoMaD outperforms ViNT by 25\% in terms of both efficiency and collision avoidance, making it the state-of-the-art approach to monocular navigation.

While NoMaD is able to learn and demonstrate impressive navigation and exploration behaviors across diverse settings, its primary limitation is that the system does not reason explicitly about the environment scale. The action candidates from NoMaD's diffusion policy are not metric; while a scaling factor can be tuned for a specific robot/camera pair, the policy is not guaranteed to produce spatially sensible actions. This can lead to collisions on robots and in environments outside of the training distribution, as seen in our evaluation (Sec. \ref{sec:comparison}, Fig. \ref{fig:NoMaD}), where we discuss the advantages and limitations of NoMaD. It should be noted that all of the aforementioned methods are too computationally heavy to run onboard a 37~g MAV and thus require offboard computation. 

In this work, we ask the following question: using \emph{only} a monocular camera for sensing obstacles, can we obtain depth maps with sufficient metric accuracy to enable 3D reconstruction of the MAV's environment? This would enable the use of motion planning and navigation techniques used by larger state-of-the-art UAVs \cite{tang2018autonomous,loquercio2021learning}. We hypothesize that a modular pipeline consisting of depth estimation, local mapping via fusion, and planning will enable significantly faster flight and more robust generalization to unseen environments \cite{gervet2023navigating}. In addition, a modular approach allows one to directly leverage improvements in depth estimation and motion planning without having to retrain an end-to-end policy from scratch. Finally, such an approach affords the ability to easily incorporate new objectives into the navigation stack (e.g., tracking a target object or constraining the drone's camera angle for cinematic applications). 

\begin{figure}[t]
\begin{center}
\includegraphics[width=\linewidth]{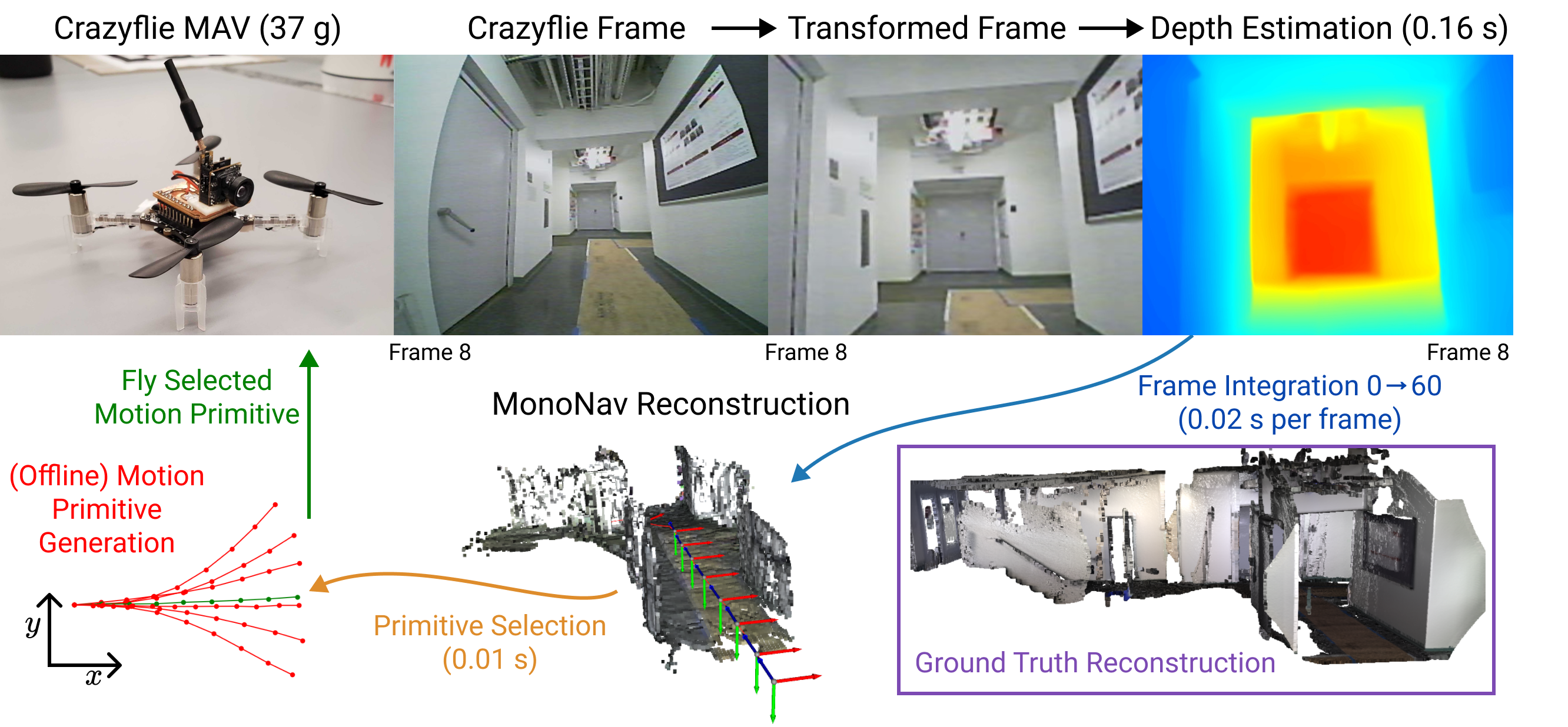}
\end{center}
\caption{MonoNav converts a series of RGB images into depth estimates (top row, left to right), then fuses them into a 3D reconstruction (bottom middle). MonoNav then selects from collision-free motion primitives (bottom left) to navigate to a goal position.}
\label{fig:mononav}
\end{figure}

\subsection{Statement of Contributions}
Our primary contribution in this work is to demonstrate that, surprisingly, a monocular system combined with state-of-the-art depth estimation techniques can perform local 3D reconstruction with sufficient quality to enable fast MAV navigation in unknown environments. We present MonoNav: a navigation stack that leverages pre-trained transformer-based models for monocular depth estimation \cite{bhat2023zoedepth,sayed2022simplerecon,ranftl2020towards} in combination with off-the-shelf fusion and planning techniques. To enable the use of pre-trained models without any fine-tuning, we propose an image processing pipeline that minimizes domain shift by performing lightweight image transformations. We perform experiments demonstrating that our framework enables the Crazyflie (a 37~g MAV) to navigate fast (0.5~m/s) in cluttered indoor environments with fully onboard sensing and offboard computation. To our knowledge, MonoNav is the first monocular navigation stack that creates a 3D metric reconstruction for navigation, with validation in hardware.

\section{Technical Approach}

MonoNav is composed of simultaneous reconstruction and planning processes (Fig.~\ref{fig:mononav}), which we describe below. Due to the limited compute onboard a MAV, MonoNav (and all state-of-the-art approaches) require offboard compute.

\subsection{Monocular depth estimation and mapping} 
The monocular mapping process is broken into two stages: metric depth estimation and fusion. We leverage recent advances in monocular depth estimation \cite{bhat2023zoedepth} that produce metric depth estimates from a single image. One of the key requirements of our navigation stack to demonstrate generalization is to use only \emph{pre-trained} models for depth estimation without any fine-tuning. Due to the domain shift that arises from the difference between the MAV's fish-eye camera and the camera used for training the depth estimation models, we propose a lightweight image pre-processing step (Fig.~\ref{fig:mononav} - top center) that transforms a source image from the MAV's camera to a target image that appears \emph{as though} it was taken with the camera used to train the pre-trained models. This can be achieved with a standard image processing library since both cameras' intrinsics are known. We then pass the transformed image to a pre-trained model which performs monocular depth estimation. 

In contrast to prior techniques that only produce depth images up to an unknown scaling factor, recent approaches \cite{bhat2023zoedepth,sayed2022simplerecon} produce absolute estimates of depth for every pixel. MonoNav uses ZoeDepth \cite{bhat2023zoedepth} (specifically, \texttt{ZoeD\_N}) for per-frame metric depth estimation. Combined with the drone's pose estimates from optical flow odometry, MonoNav uses off-the-shelf depth fusion \cite{newcombe2011kinectfusion} to create a Truncated Signed Distance Function (TSDF) representation of the environment. We represent the 3D map using Open3D's VoxelBlockGrid representation
\cite{dong2022ash}, which discretizes the world into voxels (each with a TSDF value and weight). We perform TSDF fusion on each collected depth image to construct a local map. This fusion process corrects for per-frame errors in depth estimation and also provides a memory of previously seen portions of the environment.

\subsection{Navigation} At each timestep, the robot has access to the map in the form of a VoxelBlockGrid. For collision-free navigation towards a goal, we use motion primitives; in principle, other planning approaches (e.g., A*, RRT*) could work.

The motion primitives and open-loop velocity setpoints are generated in a single offline step and stored in a trajectory library. From a desired constant speed $V$, maximum yaw rate $A$, and horizon $T$, we define our motion primitives from a Dubins' car dynamics model, with forward velocity $\dot{x}_{\text{sp}}(t) = V$ and yaw rate $\dot{\psi}_{\text{sp}}(t) = A \sin{(\pi t/T)}$. This ensures that yaw rates are zero at the beginning and end of each primitive for smooth transitions between primitives. We integrate the inputs to determine the spatial trajectory used in primitive selection. By varying $A$, we generate our library of primitives.

At runtime, the robot considers the set $\mathcal{T}$ of available trajectories $\tau \in \mathbb{R}^{n\times3}$, each consisting of $n$ position waypoints.
We also define the set $\mathcal{V}_o$ of occupied voxel coordinates $v_o \in \mathbb{R}^3$ in the VoxelBlockGrid, as well as the minimum distance $D(\tau,x)$ from any point along the trajectory $\tau$ to a coordinate $x\in \mathbb{R}^{3}$:
\begin{equation}
    D(\tau,x) = \min_{0 \leq i < n} ||\tau_i - x||_2.
\end{equation}
At each navigation step, we select the motion primitive $\tau^*$ that brings us closest to the goal position $x_g \in \mathbb{R}^3$ while maintaining a tunable minimal distance $c \in \mathbb{R}_{>0}$ from any obstacle:
\begin{align}
    \tau^* = \arg\min_{\tau\in\mathcal{T}} D(\tau,x_g) \quad
    \text{subject to}\quad  D(\tau,v_o) \geq c, \quad \forall v_o \in \mathcal{V}_o. \label{eq:primitive_selection}
\end{align}
In practice, we determine the set $\mathcal{V}_o$ by filtering all voxels in the VoxelBlockGrid by thresholds for weight, height, and TSDF value. We exhaustively compute the distances from all trajectory points to the goal position and to every occupied voxel. If no motion primitive satisfies the distance threshold criterion (i.e., Eq. \ref{eq:primitive_selection} is infeasible), the MAV is instructed to stop and land. The parameter $c$ can be decreased to increase feasibility, though the MAV may fly closer to obstacles. In this way, $c$ can be used to tune how conservatively MonoNav behaves. This self-arresting capability distinguishes MonoNav from state-of-the-art approaches like NoMaD, whose termination conditions are ``reach goal" or ``crash".

\section{Hardware Evaluations}
We implement MonoNav on the Crazyflie 2.1, a MAV configured as in \cite{kang2019generalization,mae345}. The Crazyflie is outfitted with a Flow deck v2 for position and velocity estimation and a Wolfwhoop WT05 RGB camera. Our offboard computer, which has a GeForce RTX 4090 GPU, communicates with the MAV and receives the analog video stream over radio. The Wolfwhoop camera suffers from significant `barrel distortion' due to its fish-eye lens; we transform this image to the desired camera intrinics using OpenCV's undistortion and warp affine functions.

\subsection{Depth Estimation Evaluation}

\begin{figure}[t]
\begin{center}
\includegraphics[width=\linewidth]{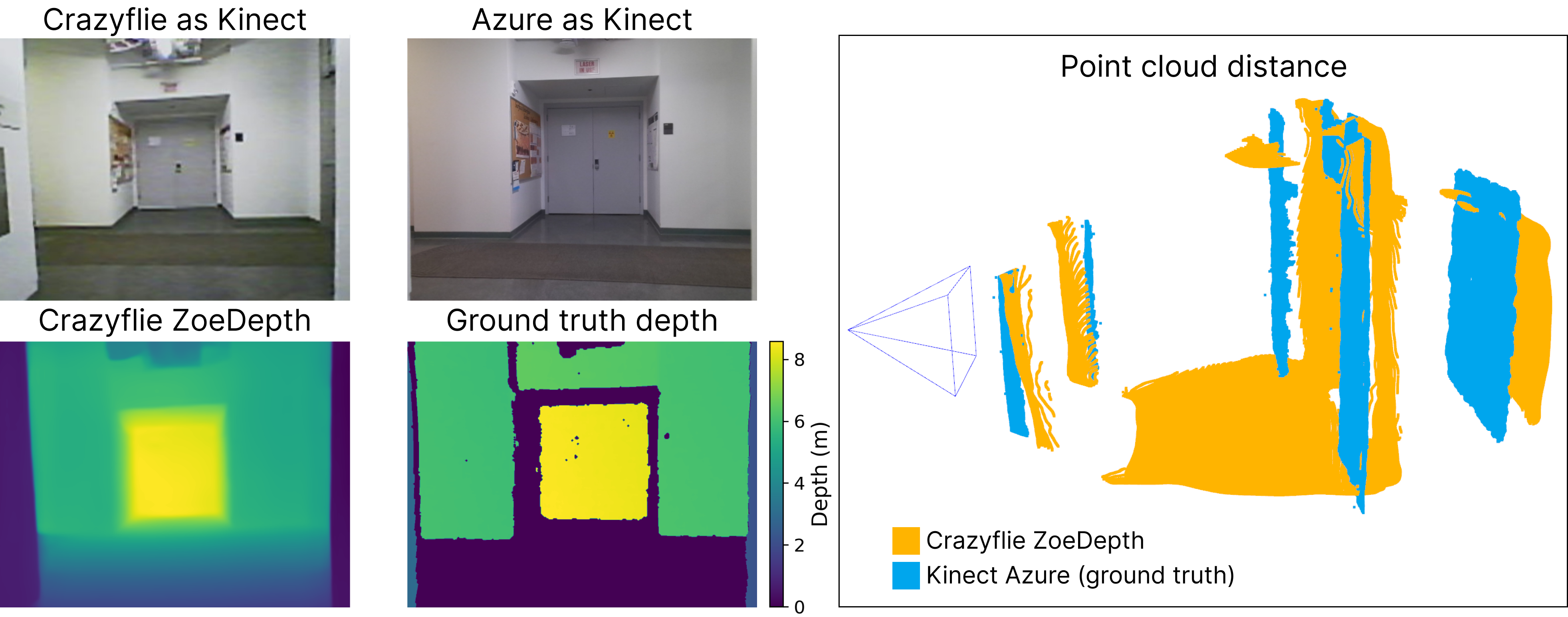}
\end{center}
\caption{(Left) After transforming the Crazyflie and Kinect Azure cameras to common intrinsics, we perform per-frame, per-pixel depth estimation evaluation. Note that the Kinect Azure filters out many points (e.g., the floor) which are represented as zeros and omitted from evaluation. (Right) To overcome per-pixel mismatch between the different cameras, we also compute the point cloud distance (PCD); i.e., for each point in the ground truth (blue) point cloud, we calculate its distance to the nearest point in our estimated (yellow) point cloud. For quantitative comparison see Table \ref{tab:depth_evaluation}.}
\label{fig:depth_evaluation}
\end{figure}

For per-frame depth estimation evaluation, we rigidly connected the Crazyflie's Wolfwhoop and Microsoft Kinect (ground truth) cameras and maneuver them along typical trajectories in indoor hallway scenes. We follow the typical approach of pixel-wise comparison \cite{eigen2014depth}; to address the differing camera intrinsics, we re-project the points to match as closely as possible. It is important to note that despite calibration, undistortion, and finding homographies to align features, the pixels do not match perfectly (see Fig. \ref{fig:depth_evaluation}), which increases the pixel-wise error. To address this, we also compare the distance between ground truth and estimated point clouds.

We follow the evaluation from \cite{bhat2023zoedepth} and determine the absolute relative error (REL) $= \frac{1}{M}\sum_{i=1}^{\put(0,-3){\scriptsize{$M$}}}|d_i-\hat{d}_i|/d_i$, the root mean squared error (RMSE) $= [\frac{1}{M}\sum_{i=1}^{\put(0,-3){\scriptsize{$M$}}}|d_i-\hat{d}_i|^{2}]^\frac{1}{2}$, the average $\log_{10}$ error $= \frac{1}{M}\sum_{i=1}^{\put(0,-3){\scriptsize{$M$}}}|\log_{10}{d_i}-\log_{10}{\hat{d}_i}|$, and the threshold accuracy $\delta_n = \%$ of pixels s.t.~$\max{(d_i/\hat{d}_i,\hat{d}_i/d_i)} < 1.25^n$ for $n=1,2,3$ (i.e., the fraction of pixels within a scale factor of $1.25^n$). The quantities $d_i$ and $\hat{d}_i$ refer respectively to the ground truth and predicted depth at pixel $i$, and $M$ is the total number of pixels in the image. In addition, to address any pixel-wise overestimation of error, we also determine the point cloud distance (PCD) $=\frac{1}{|G|}\sum_{g\in G} \min_{e \in E}||g-e||_2$; i.e., for each point $g$ in the ground-truth point cloud $G$, we calculate its distance to the closest point in the estimated point cloud $E$.
See Fig. \ref{fig:depth_evaluation} for sample images of both processes, and Table \ref{tab:depth_evaluation} for the quantitative results.

ZoeDepth's performance in the MonoNav pipeline against Kinect Azure ground truth depth is shown in Table \ref{tab:depth_evaluation}. The errors are averaged over 77 frames from a typical navigation sequence. Due to pixel mismatch between the different cameras, point cloud distance (PCD) is bolded as the fairest metric. With RMSE error of 1.05~m, and PCD error of 0.41~m, ZoeDepth is able to provide a sufficiently accurate metric depth for indoor reconstruction and navigation.

\begin{table}
\centering
\footnotesize
\begin{tabularx}{\linewidth}{@{}l|*{7}{X}|c@{}}
\toprule
\textbf{Method} &
\textbf{$\delta_1$}$\uparrow$       & \textbf{$\delta_2$}$\uparrow$          & \textbf{$\delta_3$}~$\uparrow$            & REL~$\downarrow$          & RMSE~$\downarrow$  & $\log_{10}$~$\downarrow$ & PCD~$\downarrow$\\ \midrule
MonoNav   & 0.62          & 0.85          & 0.95          & 0.48            & 1.05          & 0.11          & \textbf{0.41}             \\  
\bottomrule
\end{tabularx}
\vspace{6pt}
\caption{ZoeDepth depth estimation evaluation in the MonoNav pipeline (i.e., on a MAV camera in hallway environments). \textit{Units: meters.} Arrows indicate the direction of better performance. PCD is bolded as the fairest metric.}
\label{tab:depth_evaluation}
\end{table}

\subsection{Results: Navigation}

For hardware experiments, we define a set of motion primitives (Fig. \ref{fig:mononav}, bottom left) by $T = 1.0$~s,  $V=0.5$~m/s and $A\in\{-0.7 + k0.2\bar{3}\}_{k=0}^{6}$~rad/s (i.e., 7 evenly spaced values between [-0.7, 0.7]~rad/s). We set the distance threshold $c = 0.5$~m and the goal position $x_g = (10, 5, 0.4)$~m (in an East-North-Up world frame). The camera has a measured lag of 0.12~s, per-frame depth estimation with ZoeDepth takes 0.11-0.16~s, fusion takes 0.02~s, and motion primitive selection takes 0.01~s. Camera readings, depth estimation, and integration occur at 3-4~Hz and replanning occurs at 1~Hz. It should be noted that both fusion and planning take longer as more voxels are added to the map.

\begin{figure}[!t]
\begin{center}
\includegraphics[width=\linewidth]{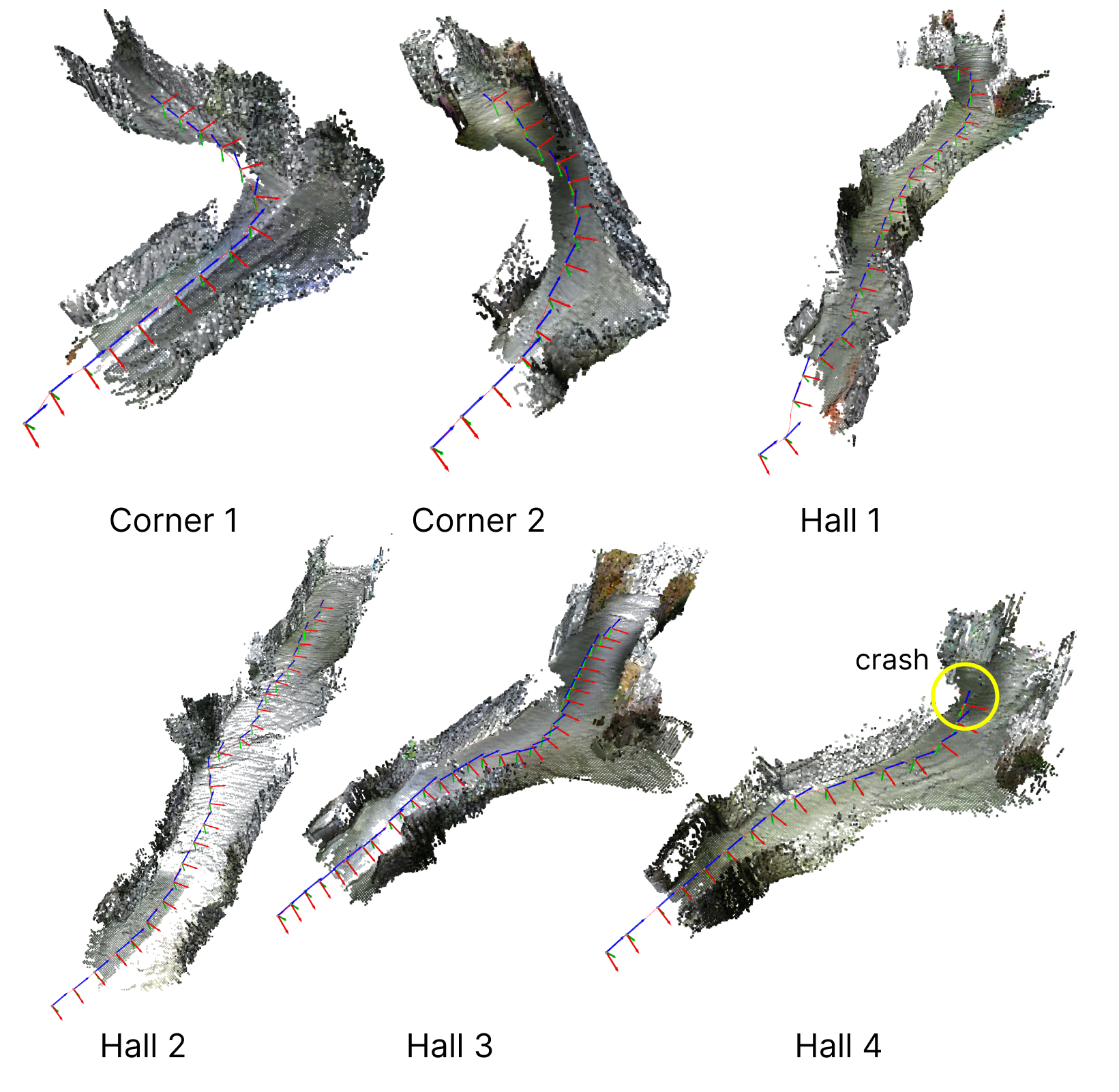}
\end{center}
\caption{MonoNav produces metric reconstructions for planning real-time, enabling navigation and collision avoidance indoors. Six representative scenes are shown above, including corners and longer hallways. Coordinate frames represent the drone's pose as flown through the trajectory.}
\label{fig:reconstructions}
\end{figure}

We test MonoNav in constrained hallway settings. These settings vary in complexity, ranging from straight sections, T-intersections, curved walls, and open spaces with columns. In 15 runs across 10 unique indoor settings (six of which are shown in Fig. \ref{fig:reconstructions}), MonoNav navigates successfully and avoids most obstacles. Of the 15 runs, MonoNav crashed once (Fig. \ref{fig:reconstructions} Hall 4), and was prematurely terminated once (Fig. \ref{fig:reconstructions} Hall 1). In both cases, MonoNav turned into a wall or dead-end that was previously occluded and thus not perceived as an obstacle. The goal position $x_g = (10, 5, 0.4)$~m induced a leftward bias into the navigation, which is reflected in the trajectories.

\section{Comparison to State of the Art: NoMaD}
\label{sec:comparison}

\begin{figure}[!b]
\begin{center}
\includegraphics[width=\linewidth]{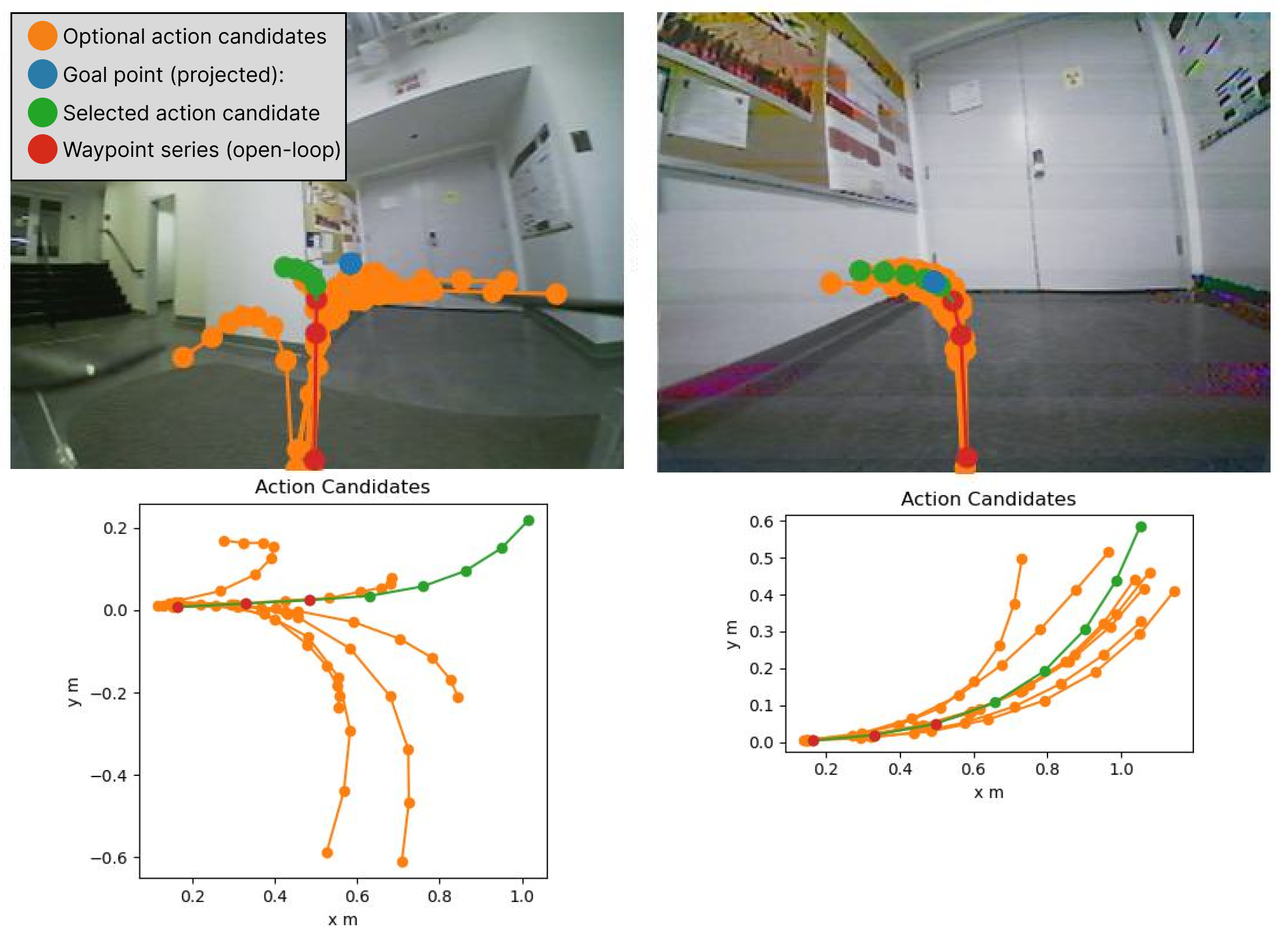}
\end{center}
\caption{\textbf{Sample action candidates from NoMaD.} The 8 action candidates are shown in orange, the goal position $x_g = (10,5)$ is projected in blue, and the selected candidate (which makes the most progress towards the goal) is in green. Open-loop navigation through the red waypoints is completed before re-planning. (Left) NoMaD demonstrates diverse, multimodal actions - avoiding the wall by turning right or left. (Right) NoMaD suggests a turn left into an obstacle.}
\label{fig:NoMaD}
\end{figure}

To evaluate MonoNav against state of the art monocular navigation techniques, we compare it to NoMaD: \textit{Goal Masked Diffusion Policies for Unified Navigation and Exploration} \cite{sridhar2023nomad}. NoMaD uses EfficientNet encoders and a Transformer decoder to transform a series of recent observations and (optional) goal image into a context. NoMaD uses the context to condition action diffusion, producing normalized action candidates which are scaled based on the robot's physical characteristics. The goal masking ensures NoMaD can operate in goal-image-directed (``navigation") and goal-image-agnostic (``exploration") modes. For the purposes of our evaluation, we only ran NoMaD in exploration mode to match MonoNav. Example action candidates are shown in Fig. \ref{fig:NoMaD}.

We evaluate NoMaD and MonoNav side-by-side in 5 unique environments. Each environment has a goal position, which encourages a certain behavior (e.g., straight, left turn, right turn). We run three trials for each method in each environment (30 runs total). We calculate performance both in terms of goal completion (\%) and collision rate, and report values in Table~\ref{tab:performance_evaluation}.  Goal completion (\% to Goal) is calculated as $1-||x_T-x_g||/||x_0-x_g||$, where $x_0,x_T,x_g$ are the initial, final, and goal positions. Collision rate is the ratio of collisions to runs.

Since all of the action candidates suggested by NoMaD should in principle be collision-free (and there is no other way to reason about proximity to obstacles), we choose the action candidate which makes the most progress towards the goal. Additionally, other than reaching the goal or manual termination, there are no other criteria for self-stopping in NoMaD as there are in MonoNav.  

For the evaluation, MonoNav has access to 11 one-second motion primitives at $V=0.5$~m/s, with $\dot{\psi}$ amplitudes defined by $A\in\{-0.7 + k0.14\}_{k=0}^{10}$~rad/s (i.e., 10 evenly spaced values between [-0.7, 0.7]~rad/s). We set the distance threshold $c = 0.2$~m. MonoNav flies each motion primitive open-loop using velocity control, resulting in smooth, chained primitives.

NoMaD accepts a series of images directly from the Wolfwhoop camera. We keep the settings identical to the original paper wherever possible, and configure NoMaD to output 8 action candidates. These action candidates are not exactly metric; through testing, a factor of 1/7 was determined to be an appropriate, conservative approach for indoor hallways. Following the paper, we follow the first 3 waypoints in open-loop fashion before re-planning.

\subsection{Results: Baseline Comparison}

\begin{figure}[!b]
\begin{center}
\includegraphics[width=\linewidth]{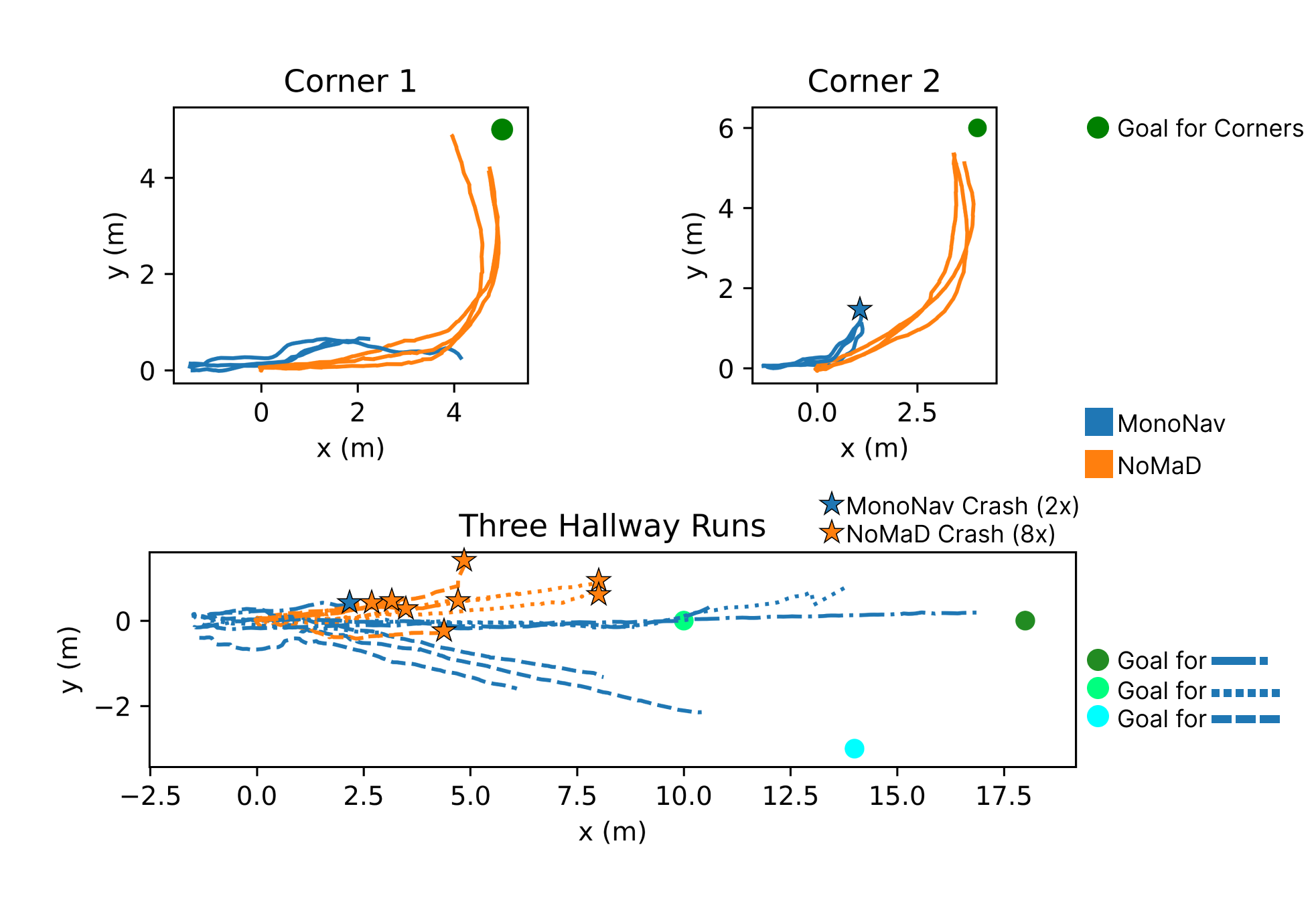}
\end{center}
\caption{We plot the trajectories of all 15 trials in 5 unique environments, the goal positions, and the crash locations. As shown in the graph, MonoNav outperforms NoMaD during straighter segments; NoMaD outperforms MonoNav in cornering, when a clear, agile maneuver is required. Note while the walls are not depicted, the hallway width is typically 2.5 meters throughout. Due to its reliance on past frames for reconstruction (see Sec. \ref{subsec:limitations}), MonoNav is ``warm-started" at $x_0=(-1.5,0.0)$ m.} 
\label{fig:results}
\end{figure}

The performance of MonoNav and NoMaD, averaged over 15 trials (each) in 5 diverse settings, is shown in Table \ref{tab:performance_evaluation} and Fig. \ref{fig:results}. We find that while MonoNav has a 22\% decrease in goal-seeking performance, it has a $4\times$ improvement in collision avoidance. This is because MonoNav can use the 3D reconstruction to reason about collisions, and stop itself if no primitive remains sufficiently far from obstacles. Noise in the state and depth estimates translates to noise in the point cloud, so MonoNav is typically over-conservative, resulting in the 22\% degradation in performance.

\begin{table}
\centering
\footnotesize
\newcolumntype{C}{>{\centering\arraybackslash}X} 
\begin{tabularx}{0.6\linewidth}{@{}l|C|C|c@{}}
\toprule
\textbf{Method} & \% to Goal $\uparrow$ & Collision Rate $\downarrow$ \\ \midrule
MonoNav   & 47.4\%          & \textbf{0.13} \\
NoMaD     & \textbf{61.0\%}          & 0.53 \\  
\bottomrule
\end{tabularx}
\vspace{6pt}
\caption{Average monocular navigation performance in 15 trials (each) across 5 environments. \% to Goal is the ratio of progress to the goal, and collision rate is the ratio of collisions. Arrows indicate the direction of best performance, and the top performer in each column is bolded.}
\label{tab:performance_evaluation}
\end{table}

\subsection{Discussion}

NoMaD is a powerful monocular navigation platform with impressive demonstrations on numerous embodiments. Additionally, it accepts fish-eye images as input, which contain much more peripheral information about obstacles than the transformed camera images do for MonoNav.

There are additional advantages as compared to MonoNav. NoMaD is more straightforward to run out of the box (e.g., no additional planner is required). Additionally, since MonoNav fuses all frames into the reconstruction ($0\rightarrow100+$), the reconstruction can be too sparse in the beginning, and too noisy by the end (resulting in premature termination). NoMaD, by comparison, requires fewer frames to produce meaningful candidates, and is able to handle dynamic environments better than MonoNav.

The main disadvantage to NoMaD is the lack of concrete spatial grounding. While NoMaD regularly produces meaningful action candidates, it is also prone to over- or under-reacting, such as a U-turn in a tight space (see Fig. \ref{fig:NoMaD}). Furthermore, since information is not preserved over longer horizons (as in MonoNav), NoMaD can be tricked (e.g., by a featureless wall or poster). In our experiments, we found that NoMaD does very well in cornering, when there is clear consensus among action candidates, but (surprisingly) poorly in straights, where the action candidates under-react and NoMaD tends to drift into walls. In these cases, NoMaD is limited by its inability to self-terminate when collision is imminent. This could be addressed by implementing a lightweight collision-detection module. Finally, the nature of the diffusion policy makes the actions stochastic and unrepeatable, making behaviors difficult to explain and reproduce.

\section{Conclusions}
We introduce MonoNav,  a navigation stack for monocular robots that generates a metric reconstruction using pre-trained depth estimation models and off-the-shelf fusion methods, enabling the use of conventional path planning techniques for collision-free navigation. We use MonoNav to enable a 37~g  MAV to navigate numerous indoor environments at 0.5~m/s. We evaluate ZoeDepth's accuracy in MonoNav, and demonstrate that it is sufficient for metric reconstruction. Finally, we compare MonoNav against a state of the art method in monocular navigation (NoMaD). Using the 3D reconstruction, MonoNav is able to reduce collision rates by a factor of 4; the trade-off of this added safety is a 22\% degradation in progress towards the goal.

\subsection{Limitations and Future Work}
\label{subsec:limitations}
MonoNav's performance is highly dependent on the quality of its reconstruction. One source of error is noise in the analog camera feed. Discolored blobs in frames, though temporary, are often integrated as occupied voxels, as are overexposed pixels when the MAV is exposed to a bright light. This can cause early self-termination as the MAV avoids a phantom obstacle. Additionally, reconstruction quality (and utility) is sensitive to errors in state estimation.

The image transformation (from fish-eye to Kinect intrinsics) narrows the field of view significantly, losing peripheral information. As a result, when MonoNav is selecting a motion primitive, the relevant depth information was seen 1.5 meters or 3 planning cycles ago. As a result, MonoNav needs to be `warm-started' with several frames before operating autonomously (e.g., in Fig. \ref{fig:results}, MonoNav starts at $(-1.5,0.0)$ m while NoMaD starts at $(0.0,0.0)$ m). In the presence of state estimation error, this reliance on previous perspectives can reduce reconstruction accuracy.

Additionally, MonoNav does not distinguish between explored and unexplored regions; unexplored regions are considered unoccupied. This architecture encourages aggressive exploration, but can lead to crashes (see Fig. \ref{fig:reconstructions} - Hall 4). A more conservative planning framework that treats space as occupied until explored could further reduce the collision rate.

In addition, active perception techniques could be used to improve reconstruction and navigation quality simultaneously. To address noise in long horizons, ``sliding window" fusion could be used instead of fusing all images. Finally, improvements in depth estimation models, state estimation, and camera will improve MonoNav's performance. Digital perception systems would vastly improve image quality over the analog feed, and products such as the HDZero Whoop Lite (7~g) may be light enough to integrate on the Crazyflie.

\newpage
\section*{Acknowledgments}
We are grateful to the authors of NoMaD \cite{sridhar2023nomad} -- specifically, Dhruv Shah -- for providing early-access to their code and model checkpoints for baseline evaluation, as well as for answering our many questions about the system. This work was partially supported by the NSF CAREER Award [\#2044149] and the Office of Naval Research [N00014-23-1-2148].

%
\small
\bibliographystyle{styles/bibtex/spmpsci_unsrt.bst}
\bibliography{iser2023}

\end{document}